\documentclass[a4paper, 11pt]{article}
\usepackage[T1]{fontenc}
\usepackage{CJK}
\usepackage{cite}
\usepackage{graphicx}
\usepackage{subfig}
\usepackage{caption}
\usepackage{tikz}
\usepackage{fancyhdr}
\usepackage{amsmath}
\usepackage[cmintegrals]{newtxmath}
\usepackage{geometry}
\usepackage{amssymb}
\usepackage{booktabs}
\usepackage{multirow}
\usepackage{textcomp}
\usepackage{enumerate}

\usepackage{cases}
\usepackage{amsmath,bm}
\usepackage{algorithmic}
\usepackage{algorithm}

\usepackage{helvet}

\title{Touch and deformation perception of soft manipulators with capacitive e-skins and deep learning} 

\author{\small{Delin Hu$^{1}$, Zhou Chen$^{1}$, Paul Baisamy$^{2}$,} \\\small{Zhe Liu$^{1}$, Francesco Giorgio-Serchi$^{2}$ and Yunjie Yang$^{1\ast}$}\\
\small{{$^{1}$SMART Group, Institute for Digital Communications,}}\\ 
\small{{The School of Engineering, The University of Edinburgh}}\\
\small{{$^{2}$The Institute for Integrated Micro and Nano Systems,}}\\ 
\small{{The School of Engineering, The University of Edinburgh}}\\
\small{{$^\ast$To whom correspondence should be addressed;}}\\
\small{{E-mail:  y.yang@ed.ac.uk}}
}
\date{}
\begin{document} 

\maketitle 

\begin{abstract}
Tactile sensing in soft robots remains particularly challenging because of the coupling between contact and deformation information which the sensor is subject to during actuation and interaction with the environment. This often results in severe interference and makes disentangling tactile sensing and geometric deformation difficult. To address this problem, this paper proposes a soft capacitive e-skin with a sparse electrode distribution and deep learning for information decoupling.  Our approach successfully separates tactile sensing from geometric deformation, enabling touch recognition on a soft pneumatic actuator subject to both internal (actuation) and external (manual handling) forces. Using a multi-layer perceptron, the proposed e-skin achieves 99.88\% accuracy in touch recognition across a range of deformations. When complemented with prior knowledge, a transformer-based architecture effectively tracks the deformation of the soft actuator. The average distance error in positional reconstruction of the manipulator is as low as 2.905$\pm$2.207 mm, even under operative conditions with different inflation states and physical contacts which lead to additional signal variations and consequently interfere with deformation tracking. These findings represent a tangible way forward in the development of e-skins that can endow soft robots with proprioception and exteroception.
\end{abstract}

\section{Introduction}
Robots are no longer limited to traditional application scenarios (e.g., large-scale industrial manufacturing), but are increasingly being employed in many aspects of our lives \cite{Dahiya:2010}, such as rehabilitation robots \cite{Yu:2015}, surgical robots \cite{Butner:2003} and social robots \cite{Belpaeme:2018}. In these emerging applications, robots need to interact with humans and environments safely and efficiently. Tactile sensing is a prerequisite to enhanced autonomy and trustworthy human-robot interaction. Flexible e-skins are one potential solution that has attracted much attention. Due to their highly deformable nature, these e-skins can improve safety during human-robot interactions and allow tactile sensors to be deployed on irregular surfaces \cite{Yang:2017}, making them available to a wide range of robotic platforms.

Over the past decades, many studies have been conducted on tactile e-skins, which rely on various transduction mechanisms \cite{Pyo:2021}, such as capacitive \cite{An:2018}, resistive \cite{Lee:2020} and piezoelectric \cite{Xie:2018} approaches. Recently, electrical resistance/impedance tomography (ER/IT) has been introduced in this area \cite{Silvera-Tawil:2015} and has shown promise in large-scale tactile sensing for rigid robots \cite{Park:2020, Chen:2021, Park:2021, Chen:2022, Park:2022}. Instead of densely deploying tactile sensing units across the region of interest (ROI), ER/IT can estimate the conductivity distribution within the ROI using only a finite set of boundary electrodes. The sparsity of electrodes is a significant advantage of ER/IT, as it effectively reduces the cost and complexity of sensor fabrication and deployment.

Soft robots have demonstrated remarkable potential in human-robot interaction in recent studies \cite{Rus:2015}. However, compared to their rigid counterparts, tactile sensing for soft robots is much more challenging. Challenges mainly arise from two aspects. First, most existing soft tactile sensors are sensitive to physical contact and body deformations \cite{Yoon:2017}. This implies that the readout signals are susceptible to interference from deformations, dramatically increasing the complexity of decoding the desired tactile information, such as contact location and force, from the signal envelope. Most previous studies only demonstrate the tactile sensing performance on undeformable substrates, where the impact of geometric deformations is not considered \cite{Park:2020, Park:2022_s, Preti:2022}. Furthermore, due to the limitations of wiring and interfaces with readout electronics, the number of soft tactile sensors that can be deployed in a unit area is restricted. Hence, methodologies that rely on sparse sensor layouts should be preferred.

This paper reports a flexible e-skin inspired by electrical capacitance tomography (ECT) \cite{Marashdeh:2015}, a type of electrical tomography similar to ERT, and learning-based e-skin data translation approaches. The e-skin consisted of a sparse liquid metal-based wirelike electrode array reduces the complexities of fabrication, wiring and interfacing compared to existing capacitive tactile e-skins \cite{Pagoli:2022vq}. By measuring capacitance between those stretchable wirelike electrodes, the e-skin can estimate the permittivity distribution in the ROI, which changes with the physical proximity and contact with external objects. Moreover, by using soft electrodes, the capacitance also changes with skin deformations. As the permittivity and geometry variations produce distinct e-skin responses, it offers the possibility to decouple the signals to obtain tactile and morphological information. 
 
The main contribution of this work is as follows.
\begin{itemize}
\item We propose and fabricate ECT-inspired e-skins with a sparse wirelike electrode layout to achieve touch recognition and deformation tracking under interference conditions. 

\item We deploy the e-skins on a soft pneumatic manipulator and characterize the sensing responses with various inflation states and physical contacts.

\item We demonstrate touch point recognition under a range of inflation states using a multi-layer perceptron (MLP) and data collected in dynamic experiments.

\item We show the effectiveness of using a transformer-based neural architecture to predict the coordinates of 5 markers deployed on the soft manipulator with different physical contacts using e-skin signals under certain prior knowledge.      
\end{itemize} 
\section{Flexible wirelike capacitive e-skin}
\subsection{E-skin design, fabrication and deployment}
\begin{figure}[b]
\centering
\includegraphics[scale=1]{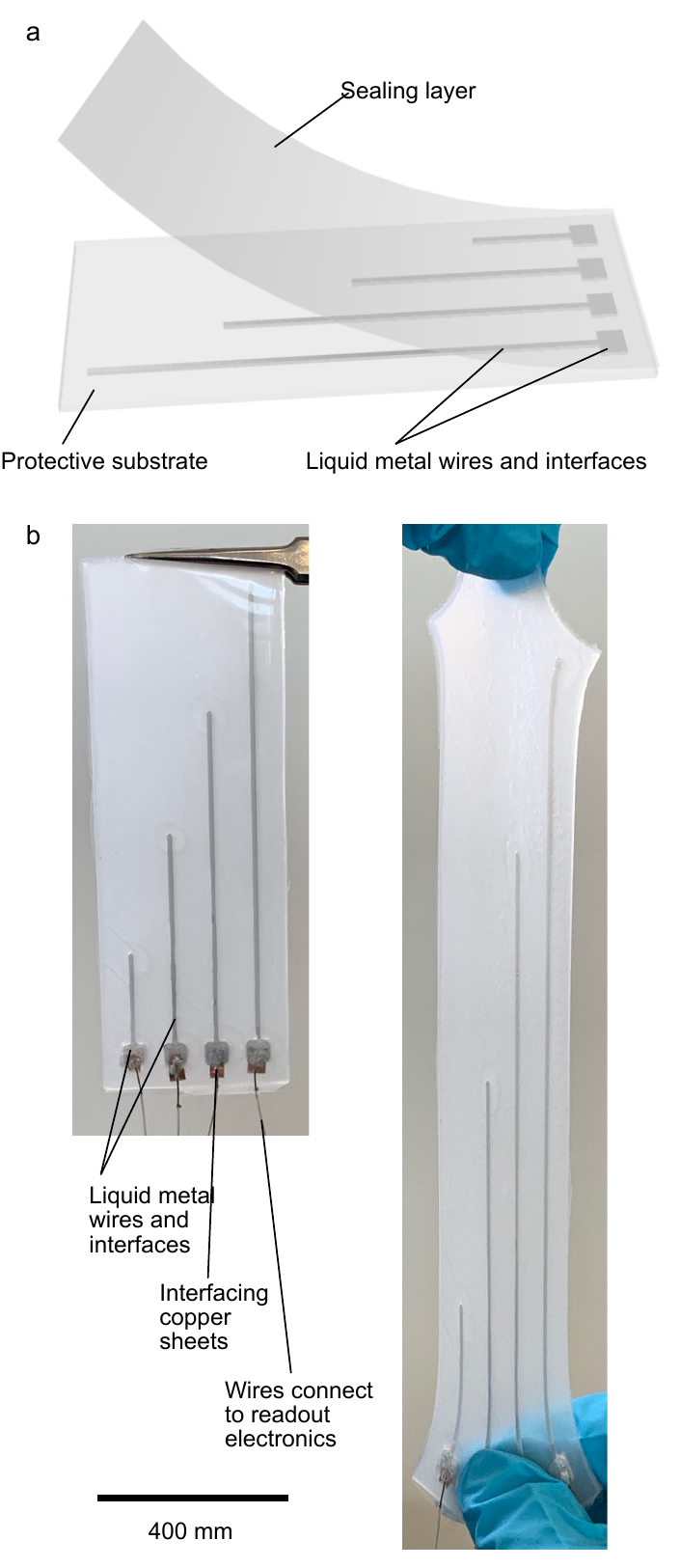}
\caption{Schematic illustration of the soft capacitive e-skin.}
\label{ECT_c4d}
\end{figure}

Figure \ref{ECT_c4d}.a provides a schematic illustration of the ECT-inspired e-skin module. ECT has been extensively studied in industrial processes, specifically for multiphase flow measurement, where the boundary electrode geometry remains constant, and the objective is to determine the permittivity distribution within the sensing region through capacitance measurements. In ECT, the capacitance between two boundary electrodes is determined by \cite{Yang:2003}
\begin{eqnarray}
\label{ECT}
C=\dfrac{Q}{V}=-\dfrac{1}{V}\iint\limits_{\Gamma}\epsilon(x,y,z)\nabla\phi(x,y,z) d\Gamma
\end{eqnarray}
where $C$ is the capacitance; $Q$ is the charge stored; $V$ is the potential difference between the boundary electrode pair; $\epsilon(x,y,z)$ and $\phi(x,y,z)$ are the permittivity and potential distribution in the sensing domain; $\Gamma$ is the electrode surface. 

When an ECT-type capacitive sensor is applied to soft robots, both the permittivity distribution and the electrode geometry are subject to change, resulting in increased complexity in obtaining desired information from capacitance data. Nevertheless, these variations can lead to distinctive patterns in capacitance measurements that can be used to extract tactile and deformation information.

Our proposed e-skin consists of 3 layers, i.e., the protective substrate, liquid metal wires and interfaces, and the sealing layer. The size of the e-skin is 40$\times$110 mm. The 4 liquid metal wires act as electrodes, and the combinations of any two electrodes form capacitors. Different lengths of wires are designed to magnify the difference in capacitance signals when touching different areas of the e-skin. The wires are 1 mm wide and 20, 45, 70 and 95 mm long, respectively. The size of the liquid metal interfaces is 5$\times$5 mm.

We fabricate the protective substrate with micro-channels for liquid metal wires and interfaces using Ecoflex 00-30 and a 3D-printed casting. A silicone membrane manufactured by film coating is bonded to the substrate as the sealing layer using uncured silicone as the adhesive. We inject Eutectic Gallium 75.5\% Indium 24.5\% (EGaIn) ink from the interfaces and exhaust the air through the ends of the wires. Fig. \ref{ECT_c4d}.b shows the snapshots of the e-skin without deformation and at 100\% elongation.  

\begin{figure}[b]
\centering
\includegraphics[scale=0.73]{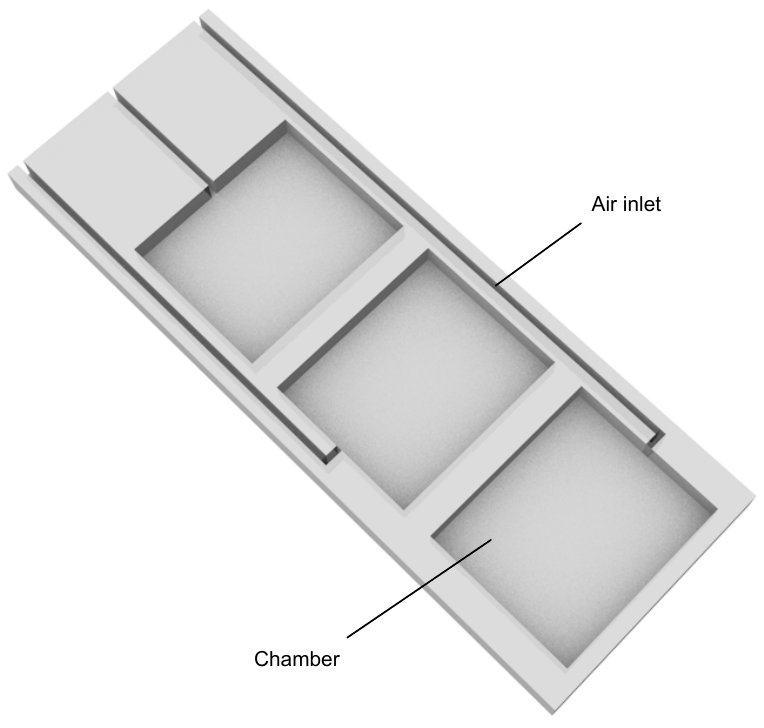}
\caption{The structure of the 3-chamber pneumatic manipulator.}
\label{robot_c4d}
\end{figure}

A 3-chamber pneumatic manipulator (50$\times$120 mm) is selected as the testbed. Pneumatic soft robots are frequently used in many applications and can provide different types of deformations, e.g., inflation and bending. The structure of the robot is shown in Fig. \ref{robot_c4d}. The size of each chamber is 40$\times$30 mm. The width of each inlet is 1.5 mm. We bond two e-skin modules to the robot surfaces (one on the front and the other on the back) using uncured silicone as the adhesive. They form an 8-electrode capacitive sensor array (each module has 4 electrodes), generating 28 independent capacitance readouts per measurement frame following the 8-electrode ECT sensing strategy. 

\subsection{E-skin Characterization}
\subsubsection{Inflation}
\begin{figure}[h]
\centering
\includegraphics[scale=0.85]{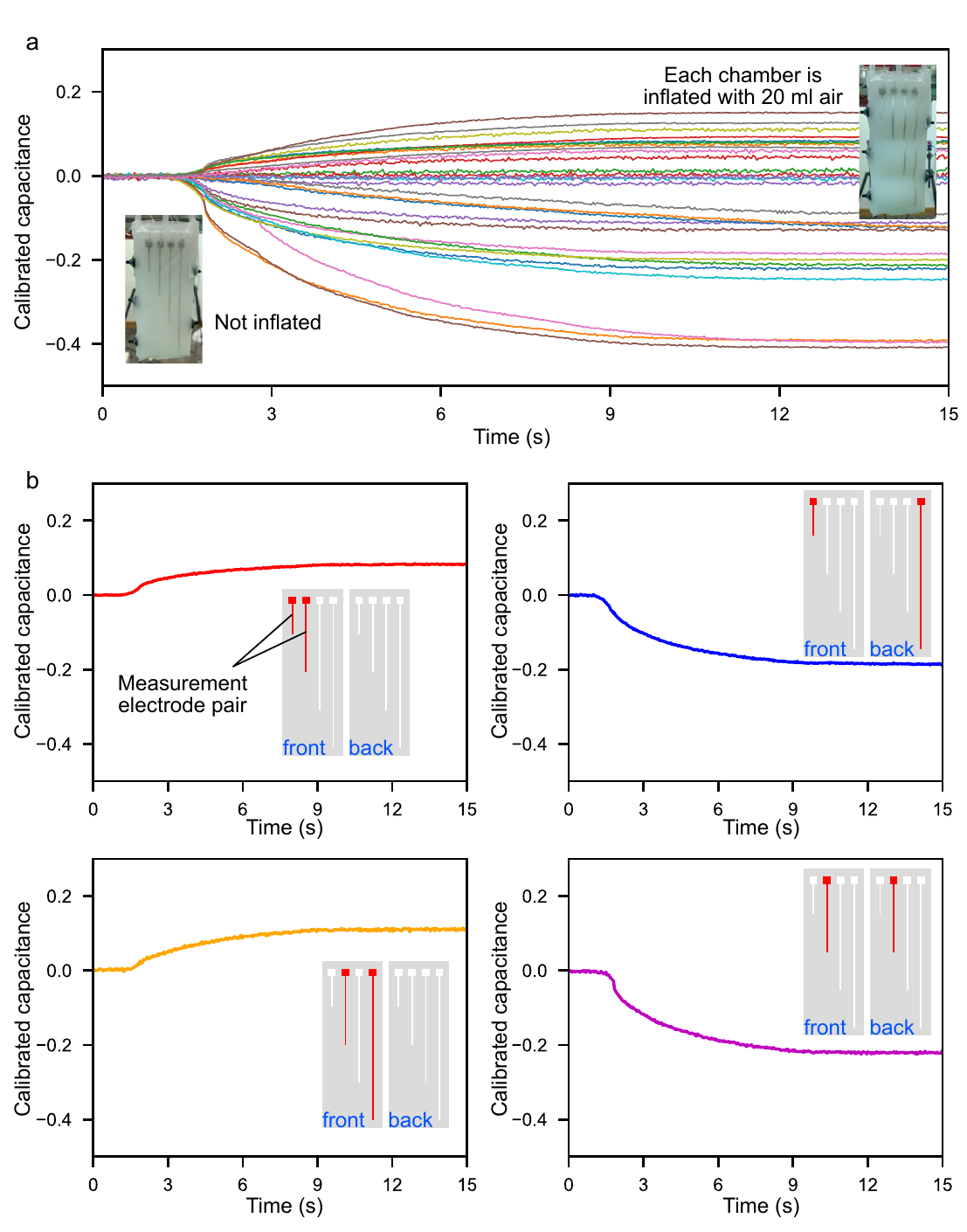}
\caption{Capacitance response of the (20,20,20) ml inflation. a, Calibrated capacitance of all 28 capacitance readouts (illustrated by different colors). b, Calibrated capacitance of four selected electrode pairs.}
\label{c_inflation}
\end{figure}

 A 3-dimensional vector $\bm{p}=(p_1,p_2,p_3)$ is used to describe the inflation state of the manipulator, where $p_1$, $p_2$ and $p_3$ are the volumes of air injected into the 1$^{st}$, 2$^{nd}$ and 3$^{rd}$ chambers, respectively. We inject 20 ml air into each chamber simultaneously, i.e., $\bm{p}=(20,20,20)$ ml, and record the e-skin response signals. Figure \ref{c_inflation}.a shows the 28 capacitance readouts during the inflation process. The y-axis is the calibrated capacitance $c$ (i.e., relative change in capacitance), which can be computed by 
\begin{eqnarray}
\label{calib}
c = \dfrac{c_t-c_0}{c_0}
\end{eqnarray}
where $c_t$ is the capacitance readout in the current state and $c_0$ is the capacitance readout in the reference state (the state without inflation).

The $\bm{p}=(20,20,20)$ ml inflation can stimulate a maximum variation of around 40\% relative capacitance change. Fig. \ref{c_inflation}.b shows capacitance readouts of four selected electrode pairs separately. The measurement electrode pairs are marked in red in the figure. On the left side of Fig. \ref{c_inflation}.b, the capacitors are formed by electrodes on the same surface and the capacitance readouts increase as the pneumatic robot is inflated. On the right side of the figure, the capacitors are formed by electrodes from different surfaces and the capacitance readouts show the opposite trend with that generated by the same-surface electrode pairs. The robot body's inflation makes the electrode area (positively correlated to capacitance) larger and the distance between electrodes (negatively correlated to capacitance) longer. For capacitors formed by electrodes on the same surface, the increase in the area of electrodes dominates the change in capacitance. For capacitors formed by electrodes on different surfaces, the variation in the distance between electrodes is more significant, which dominates the capacitance variation. 

\subsubsection{Touch under inflation}
\begin{figure}[h]
\centering
\includegraphics[scale=0.8]{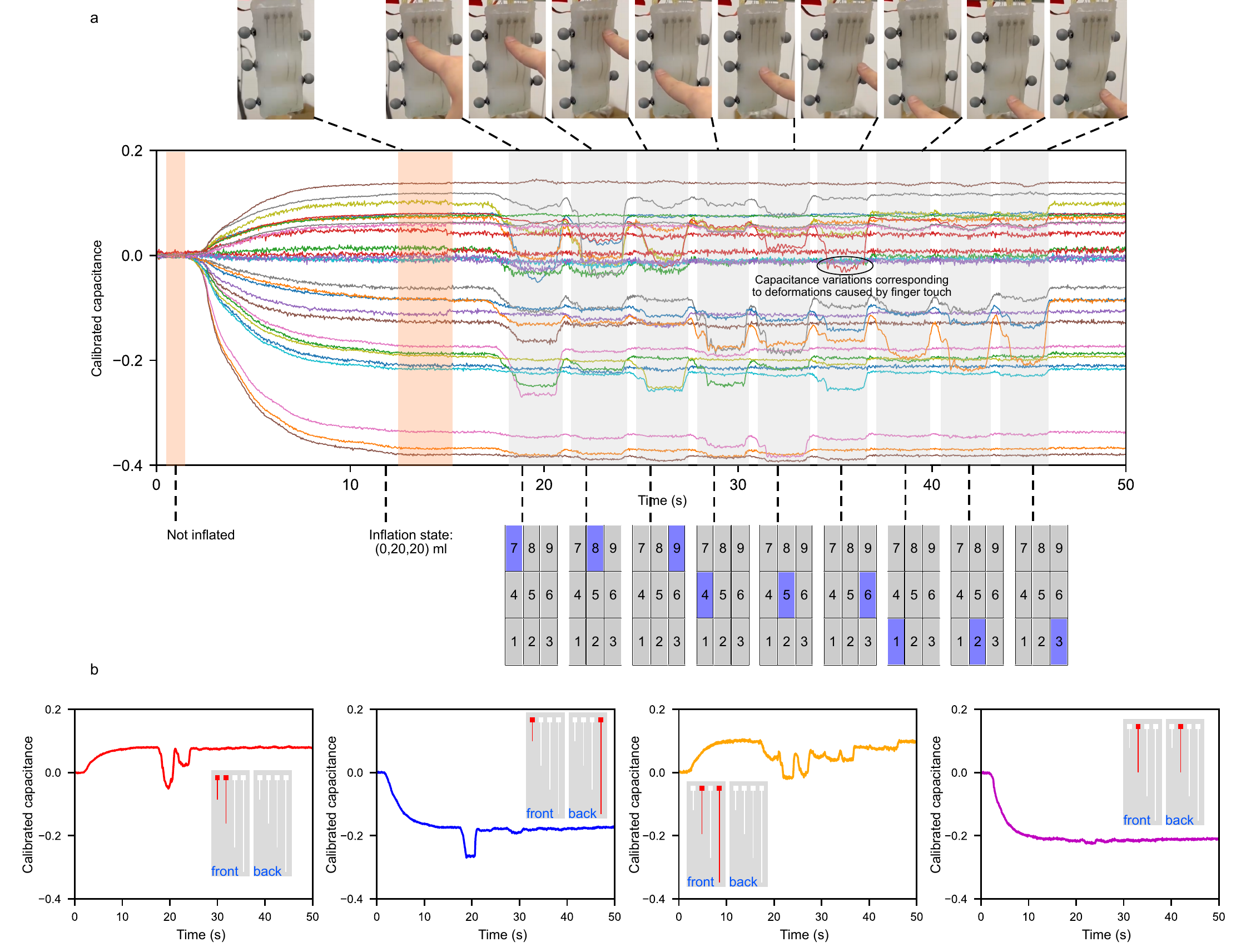}
\caption{Capacitance response of the 2-stage experiment. a, Calibrated capacitance of all 28 capacitance readouts. b, Calibrated capacitance of four selected electrode pairs.}
\label{c_touch}
\end{figure}

A 2-stage experiment is implemented to show the difference of capacitance change induced by touch (permittivity variation) and deformation (geometry variation). In the first stage, the robot is inflated to (0,20,20) ml without touch. Then we divide the front surface of the robot into 9 regions (see Fig. \ref{c_touch}.a) and touch each region individually. Fig. \ref{c_touch}.a shows the overall response of the e-skin.  

The capacitance response of the (0,20,20) ml inflation is similar to that of (20,20,20) ml inflation shown in Fig. \ref{c_inflation}. Inflation can trigger variations in every capacitance readout simultaneously, while touch only stimulates changes in a part of readouts based on the location of the touch point. Examples of capacitance readouts from 4 selected electrode pairs (marked in red) during the 2-stage experiment are shown in Fig. \ref{c_touch}.b. It illustrates that different electrode pairs have different perceptive fields. A capacitance readout only reflects the touch within its own perceptive field and is not sensitive to touches outside the area. For example, in the first graph, the capacitance readout fluctuates from 15 s to 25 s and keeps constant during other periods. It demonstrates that the capacitance readout is sensitive to the contacts on sub-regions 7 and 8, and is not affected by the contacts on the other sub-regions. This feature results in different patterns in capacitance signals induced by inflation and touch. We also observed small fluctuations in capacitance readouts during touch. This is induced by deformations of the manipulator caused by touch (e.g., bending).

\section{Experiment and data acquisition}
\begin{figure}[h]
\centering
\includegraphics[scale=1]{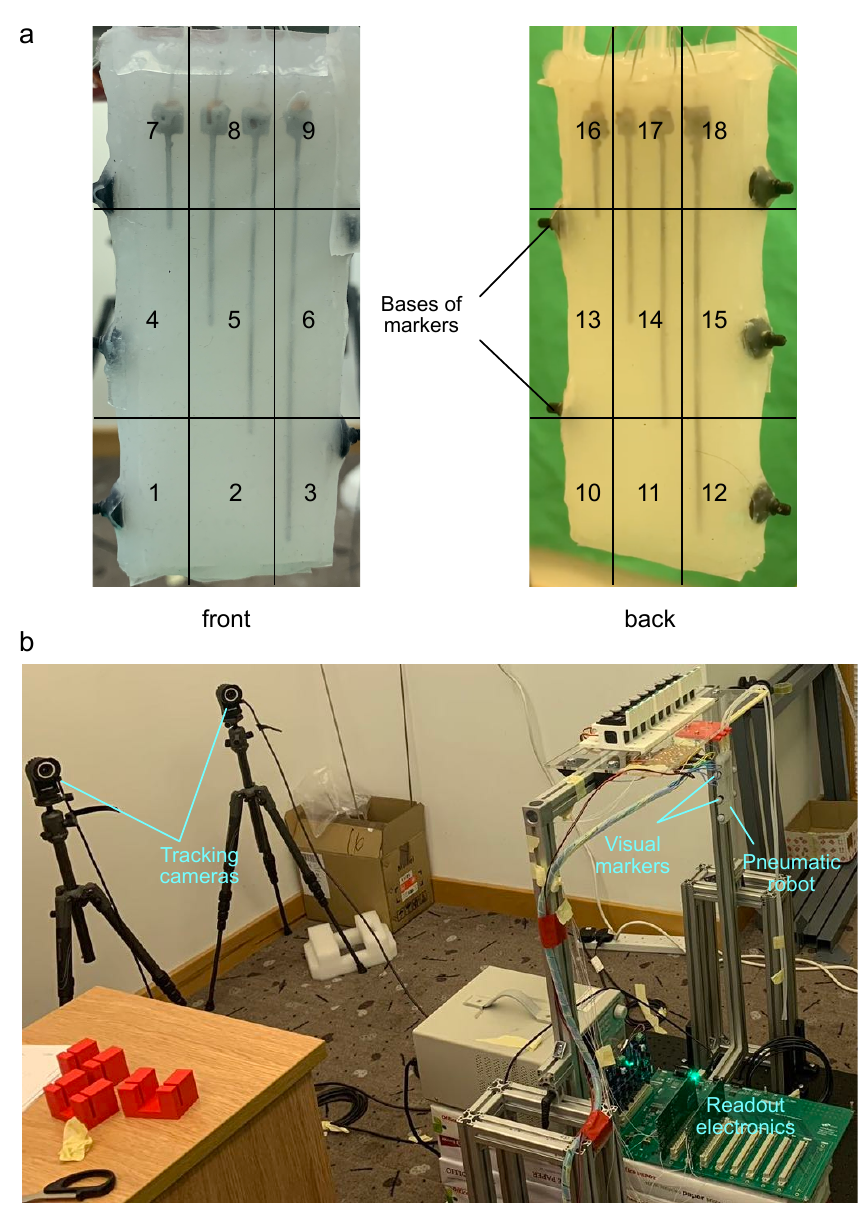}
\caption{Experiment setup. a, Division of the robot body for touch recognition. b, Experiment platform.}
\label{division}
\end{figure}

We conduct experiments to collect data to verify the feasibility of recognising touch during deformation (e.g., inflation) using the proposed flexible e-skin. Ideally, it requires us to acquire e-skin signals and the location of the contact point (as labels) simultaneously for training and verification. However, accurate touch locations are hard to pinpoint in dynamic experiments where the e-skin is not fixed on a stationary platform but installed on a manipulator that can move and deform. Therefore, we divide the surface of the manipulator into 18 sub-regions (see Fig. \ref{division}.a) and randomly subject each sub-region to touch while recording the e-skin signals. In this way, we are able to identify the index of a sub-region as a low-granularity location of the touch point. 

Furthermore, we investigate the possibility of extracting deformation information from e-skin signals, as deformation tracking is critical and sensing devices with multiple functions are desired in soft robotics. For the pneumatic robot platform, the inflation information is usually known, as the air volume injected into the chamber is controlled by actuation motors or pumps. Therefore, we focus on deformations caused by user interaction, such as bending induced by touch. To acquire deformation labels during the experiment, we bond 5 reflective visual markers on the sides of the robot and utilize three OptiTrack Flex 13 cameras to capture real-time 3D coordinates of the markers. The acquired coordinates briefly describe the deformation and serve as the deformation labels in the subsequent analysis.

Fig. \ref{division}.b shows the experimental platform. It includes the pneumatic robot equipped with the 8-electrode capacitive e-skin and 5 reflective visual markers, OptiTrack Flex cameras and readout electronics (developed in our previous work \cite{YangY:2017} that can reach 1 fF capacitance measurement resolution and over 60 dB signal to noise ratio for all measurement channels). The cameras and readout electronics data recording speed is set to 30 fps.  

The experiment includes 2 stages. In the first stage, the soft manipulator is inflated to a steady inflation state according to a preset $\bm{p}$. Each element in $\bm{p}$ can be 0 ml, 10 ml, and 20 ml, thus combining a total of 27 different inflation states. A period of 30 s capacitance data for each inflation process is recorded and is annotated as no contact. In the second stage, we randomly touch a sub-region of the robot and induce different deformations by changing the contact force. The contact and deformation process lasts 30 s. The tracking cameras and readout electronics synchronously record data during this period. We implement the same process for 6 different touch sub-regions (3 on the front and 3 on the back) in the same inflation state.  

The 2-stage experiment is repeated 27 times, corresponding to 27 different inflation states. Although only 6 of 18 sub-regions are touched in each inflation state, each sub-region is touched at least once during the experiment of the 27 different inflation states. The proposed method is expected to have a generalisation ability that enables it to infer tactile information of unseen samples. Therefore, collecting contact data of all sub-regions in one inflation state is unnecessary. 

For touch recognition, 189 groups of data are recorded (27 different inflation states, 6 touch points and 1 no contact data in each state, 27$\times$7=189). Each group of data includes 30-second capacitance signals (i.e., 900 frames at a sampling speed of 30 fps) and the index of the corresponding touch point. The data is exclusively divided into training (125$\times$30=3750 seconds, 3750$\times$30=112500 frames), validation (32$\times$30=960 seconds, 960$\times$30=28800 frames) and testing (32$\times$30=960 seconds, 960$\times$30=28800 frames) sets. 

For deformation tracking, a group of data consists of capacitance signals of 30 s and the trajectory of visual markers recorded by cameras. The data without touch and suffering from visual occlusion issues is manually filtered. Eventually, 146 groups of data are acquired, which are exclusively divided into training (108$\times$30=3240 seconds, 3240$\times$30=97200 frames), validation (18$\times$30=540 seconds, 540$\times$30=16200 frames) and testing (20$\times$30=600 seconds, 600$\times$30=18000 frames) sets. 

\section{Data analysis and results}
\subsection{Touch recognition}
\subsubsection{Neural network architecture}
\begin{figure}[t]
\centering
\includegraphics[scale=0.9]{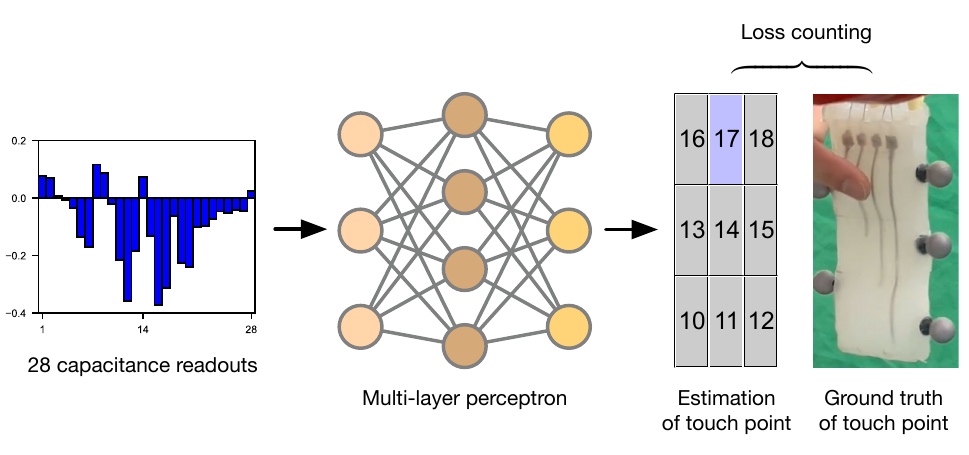}
\caption{The framework to estimate the touch point using an MLP.}
\label{neural_arch1}
\end{figure}

We employ a multi-layer perceptron (MLP) \cite{LeCun:2015} to achieve touch point classification (see Fig. \ref{neural_arch1}). There are 19 different classes in this study, 18 different touch points and one case without touch. The input of the MLP is 28 calibrated capacitance readouts in one frame. The MLP outputs a vector with a size of 19, indicating the class probability. Cross-entropy is selected as the loss function. The MLP has one hidden layer with 128 neurons. The activation function for the hidden layer is ReLU. Dropout ($p=0.1$) is used to prevent overfitting. 

\subsubsection{Training}
The training of the MLP is implemented in Pytorch. We use the Adam optimizer \cite{Kingma:2015} to update the learnable parameters to minimize the cross-entropy loss between the predicted and ground truth touch points. The initial learning rate is set to 0.001 and decayed by a factor of 1.2 every 15 epochs. We run 100 epochs of training with a batch size of 256 using the training set and save the network with the smallest loss on the validation set. The training process takes 10 minutes on one Nvidia Quadro P5000 GPU card.

\subsubsection{Results}
\begin{figure}[b]
\centering
\includegraphics[scale=0.8]{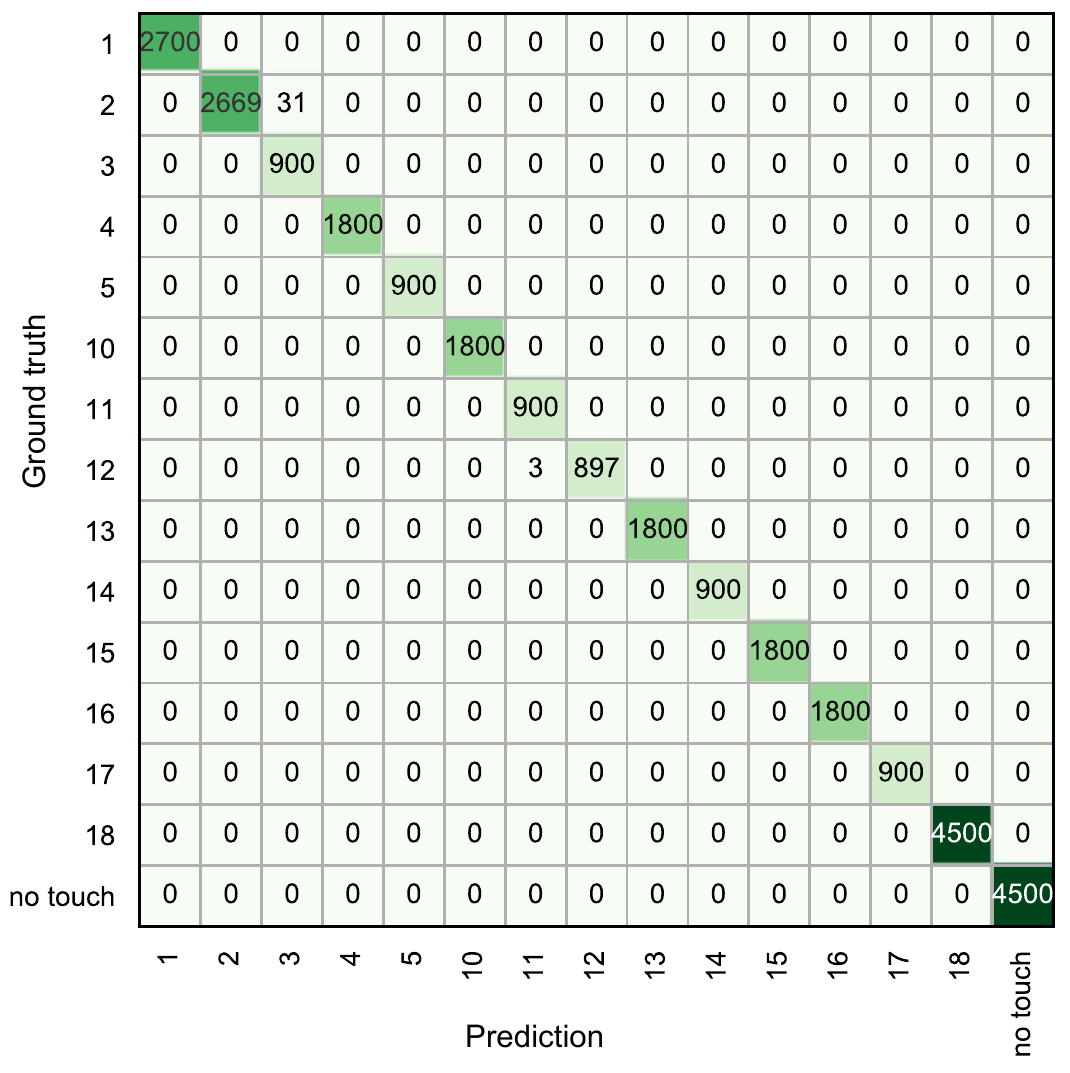}
\caption{Confusion map of the classification results.}
\label{confusion}
\end{figure}

After training, the MLP can achieve 99.88\% classification accuracy on the testing set. It demonstrates that the proposed flexible e-skin can estimate coarse touch point location using a simple learning model even when the signals are seriously interfered by the inflation of the manipulator's body. The confusion map of the classification results is shown in Fig. \ref{confusion}. It suggests that only 34 out of 28800 frames of testing samples are misclassified. All misclassifications occur between adjacent sub-regions. For example, 31 touches on sub-region 2 are incorrectly classified as sub-region 3. This is because signals induced by touches on adjacent sub-regions have relatively high similarity, which probably confuses the network.   

\subsection{Deformation tracking}
\subsubsection{Neural network architecture}
\begin{figure}[t]
\centering
\includegraphics[scale=0.85]{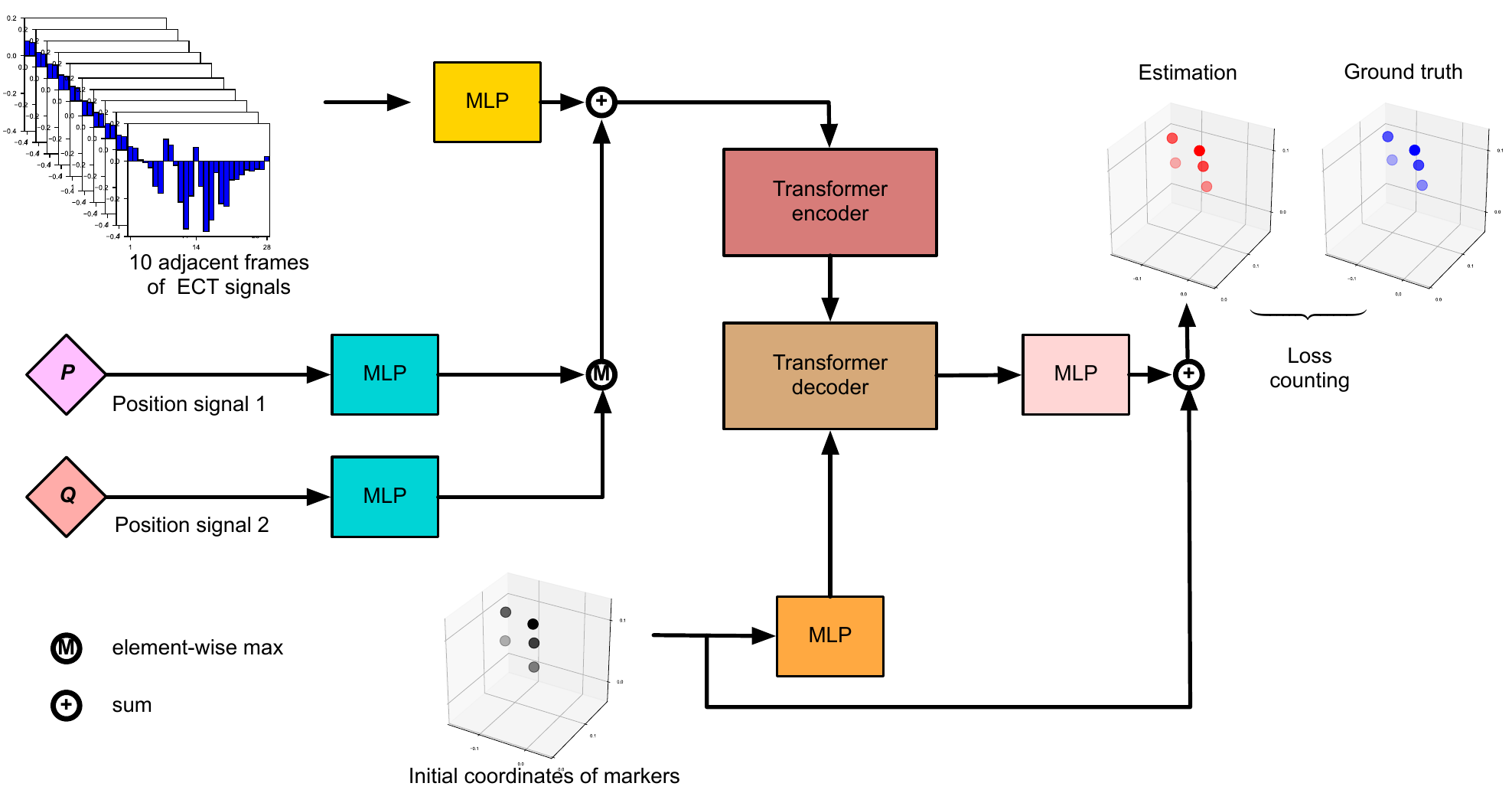}
\caption{The structure of the C2DT to estimate coordinates of visual markers based on capacitance signals. Note different colors represent different MLPs and $P, Q$ share the same MLP.}
\label{neural_arch2}
\end{figure}

Estimating coordinates of visual markers based on capacitance signals can be treated as a set-to-set issue. To this end, we apply the capacitance-to-deformation transformer (C2DT) \cite{Delin:2022}, a transformer-based architecture developed in our previous work that successfully reconstructs point clouds of the geometry of the soft robot. The structure of the C2DT is shown in Fig. \ref{neural_arch2}. The cameras and readout electronics are synchronized by an auto-click script, which leads to a slight delay between data recorded by different devices. To alleviate this problem, we input 10 frames of calibrated capacitance signals to C2DT. The position signals ($\bm{P}$, $\bm{Q}$) consist of the location information of the electrode pairs to form the capacitors, which can help the C2DT distinguish capacitance readouts generated by different electrode pairs. The squared error between estimation and ground truth is selected as the loss function.

This study focuses on deformations caused by user interactions (e.g., touch) rather than inflation (which in most scenarios is a known parameter, given that the volume of air injected into the chamber is a control parameter). The signals induced by these deformations are much smaller compared with signals triggered by inflation and touch. This makes it extremely challenging and in some cases altogether unfeasible to directly estimate the coordinates of the visual markers. In order to address this issue, we use the first frame in each trajectory as prior knowledge, i.e., the capacitance readouts are used as the reference to calibrate the capacitance input (see Eq. \ref{calib}) and the coordinates of the markers are used as the source sequence (initial coordinates of markers in Fig. \ref{neural_arch2}) of the transformer decoder.  

\subsubsection{Training}
The training of the C2DT is implemented in Pytorch. We use the Adam optimizer \cite{Kingma:2015} to update the learnable parameters to minimize the squared loss between predicted and ground truth coordinates. The initial learning rate is set to 0.001 and decayed by a factor of 1.2 every 15 epochs. We run 150 epochs of training with a batch size of 255 using the training set and save the network with the smallest loss on the validation set. The whole training process takes 2.5 hours on 3 Nvidia Quadro P5000 GPU cards.

\subsubsection{Results}
We use the average distance (AD) between the estimated and ground truth coordinates of the markers to evaluate the performance of the C2DT, which is defined as:
\begin{eqnarray}
\label{ECT}
\text{AD}=\dfrac{1}{NM}\sum_{i=1}^{N}\sum_{j=1}^{M}\sqrt{(\bm{p}_{i,j}-\hat{\bm{p}}_{i,j})^2}
\end{eqnarray}
where $N$ is the number of samples in the testing set, $M$ is the number of visual markers, $\bm{p}_{i,j}$ is the ground truth coordinates of the $j^{th}$ visual marker for the $i^{th}$ testing sample and  $\hat{\bm{p}}_{i,j}$ is the estimated coordinates of the $j^{th}$ visual marker for the $i^{th}$ testing sample. After training, the C2DT can achieve 2.905$\pm$2.207 mm AD error. This demonstrates that, with prior knowledge (the capacitance signals and coordinates of markers in the first frame of each trajectory), the proposed e-skin can be applied to track the deformation using C2DT even in operative conditions subject to severe interference (e.g. inflation and permittivity variations caused by touch).

\begin{figure}[t]
\centering
\includegraphics[scale=0.9]{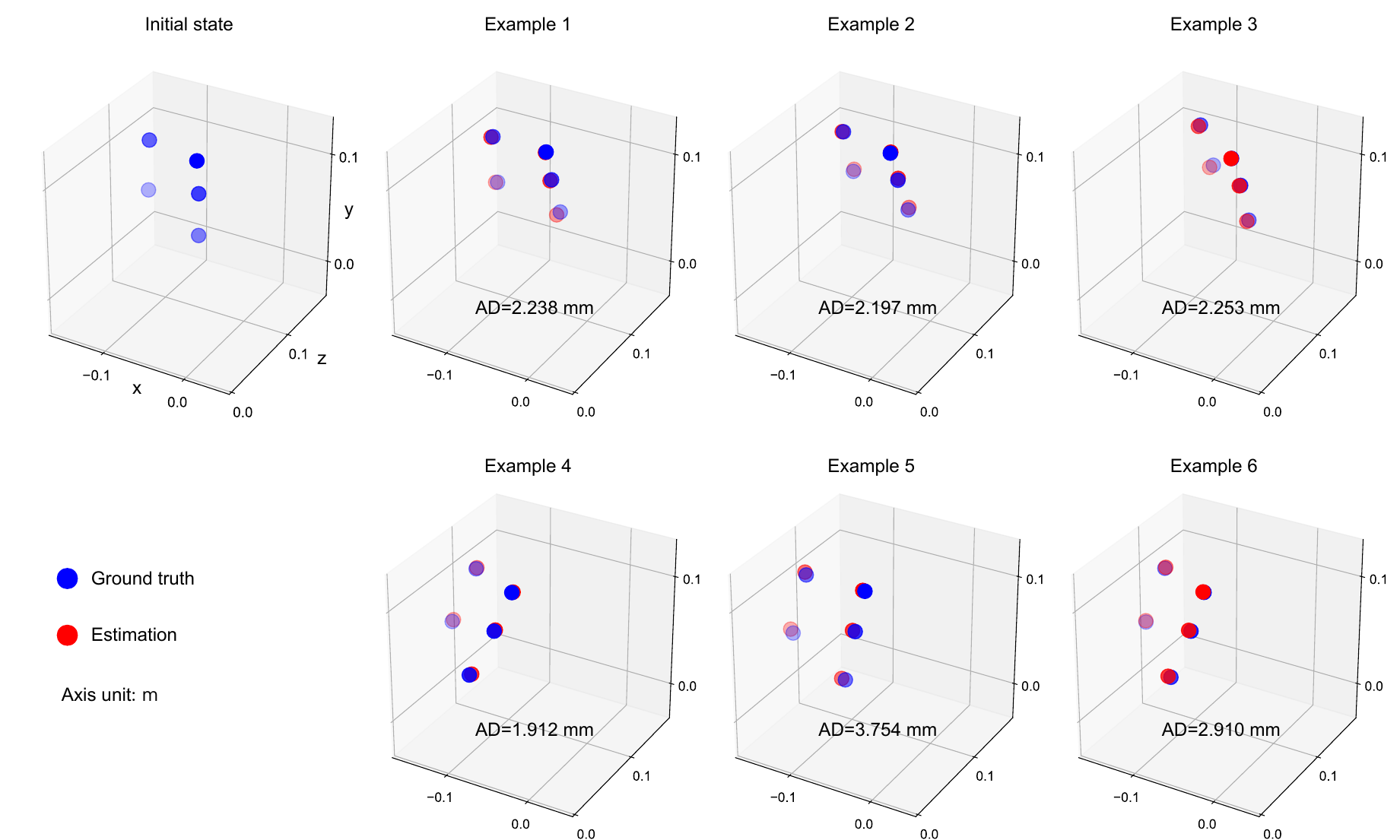}
\caption{Examples of tracking results.}
\label{tracking}
\end{figure}

\begin{figure}[h]
\centering
\includegraphics[scale=0.9]{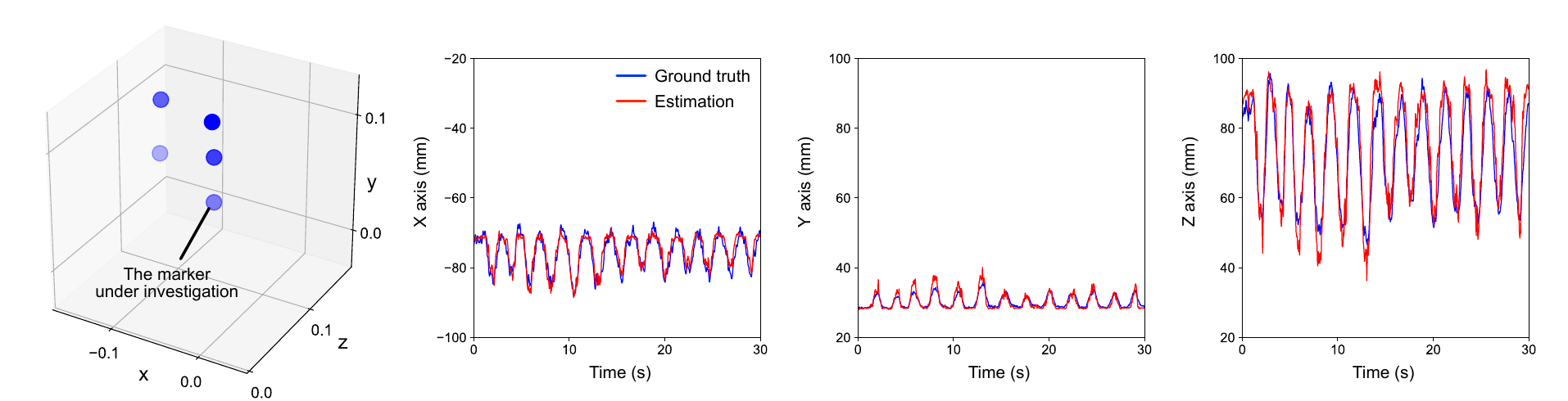}
\caption{A trajectory of tracking results of a selected marker.}
\label{tracking1}
\end{figure}

The examples of several tracking tests are shown in Fig. \ref{tracking}, where the locations of the estimated visual markers (red) and the ground truth visual markers (blue) are presented. For all cases shown, the estimated locations of the visual markers are close to the ground truth ones, indicating the high level of accuracy of the deformation tracking (maximum AD error among the 6 examples is 3.754 mm). Fig. \ref{tracking1} illustrates the tracking performance of a selected marker over a 30-s period. No significant tracking errors are observed during the whole trajectory, which further demonstrates the remarkable performance of the proposed method. 

\section{Conclusions}
In this paper, we report a flexible capacitance e-skin based on silicone and liquid metal with an ECT-inspired sparse electrode distribution. We demonstrated that the proposed e-skin can estimate contact points and track deformations on a stereotypical soft pneumatic actuator with coupling signals. Prediction of contact location under a range of actuation conditions was successfully demonstrated at a low level of spatial granularity. Furthermore, by using the first frame in a trajectory as prior knowledge and feeding it into a transformer-based architecture (C2DT), we are able to estimate the deformation of the actuator in any following frames with a high level of accuracy.  These achievements are evidence of the potential of our proposed method to be used in challenging applications, such as human-robot interaction, which typically involve various stimuli sources (e.g., deformation induced by actuation and physical contacts). In the future, multiple sensing units based on different sensing mechanisms can be integrated into the e-skin to improve accuracy and spatial resolution, enabling more complex tactile sensing, such as multiple contact points detection.

\section*{Acknowledgments}

D.H. acknowledges the support of the studentship from The University of Edinburgh. Y.Y. and F.G.S. thank the financial support of the Data Driven Innovation Chancellor's Fellowship, the Wellcome Trust iTPA fund and the MRC IAA fund.

\begin{thebibliography}{99}

\bibitem{Dahiya:2010}
R.~S. Dahiya, G.~Metta, M.~Valle, and G.~Sandini.
\newblock Tactile sensing---from humans to humanoids.
\newblock {\em IEEE Transactions on Robotics}, 26(1):1--20, 2010.

\bibitem{Yu:2015}
H.~Yu, S.~Huang, G.~Chen, Y.~Pan, and Z.~Guo.
\newblock Human--robot interaction control of rehabilitation robots with series
  elastic actuators.
\newblock {\em IEEE Transactions on Robotics}, 31(5):1089--1100, 2015.

\bibitem{Butner:2003}
S.~E. Butner and M.~Ghodoussi.
\newblock Transforming a surgical robot for human telesurgery.
\newblock {\em IEEE Transactions on Robotics and Automation}, 19(5):818--824,
  2003.

\bibitem{Belpaeme:2018}
T.~Belpaeme, J.~Kennedy, A.~Ramachandran, B.~Scassellati, and
  F.~Tanaka.
\newblock Social robots for education: A review.
\newblock {\em Science Robotics}, 3(21):eaat5954, 2022/11/07 2018.

\bibitem{Yang:2017}
Y.~Zhang, G.~Laput, and C.~Harrison.
\newblock Electrick: Low-cost touch sensing using electric field tomography.
\newblock In {\em Proceedings of the 2017 CHI Conference on Human Factors in
  Computing Systems}, CHI '17, pages 1--14, New York, NY, USA, 2017.
  Association for Computing Machinery.

\bibitem{Pyo:2021}
S.~Pyo, J.~Lee, K.~Bae, S.~Sim, and J.~Kim.
\newblock Recent progress in flexible tactile sensors for human-interactive
  systems: From sensors to advanced applications.
\newblock {\em Advanced Materials}, 33(47):2005902, 2022/11/08 2021.

\bibitem{An:2018}
B.~ An, S.~Heo, S.~Ji, F.~Bien, and J.~Park.
\newblock Transparent and flexible fingerprint sensor array with multiplexed
  detection of tactile pressure and skin temperature.
\newblock {\em Nature Communications}, 9(1):2458, 2018.

\bibitem{Lee:2020}
J.~Lee, J.~Heo, Y.~Kim, J.~Eom, H.~Jung, J.~Kim, I.~Kim, H.~Park, H.~Mo, Y.~Kim, and S.~Park.
\newblock A behavior-learned cross-reactive sensor matrix for intelligent skin
  perception.
\newblock {\em Advanced Materials}, 32(22):2000969, 2022/11/10 2020.

\bibitem{Xie:2018}
M.~Xie, Y.~Zhang, M.~Kra{\'s}ny, C.~Bowen, H.~Khanbareh,
  and N.~Gathercole.
\newblock Flexible and active self-powered pressure, shear sensors based on
  freeze casting ceramic--polymer composites.
\newblock {\em Energy \& Environmental Science}, 11(10):2919--2927, 2018.

\bibitem{Silvera-Tawil:2015}
D.~Silvera-Tawil, D.~Rye, M.~Soleimani, and M.~Velonaki.
\newblock Electrical impedance tomography for artificial sensitive robotic
  skin: A review.
\newblock {\em IEEE Sensors Journal}, 15(4):2001--2016, 2015.

\bibitem{Park:2020}
K.~Park, H.~Park, H.~Lee, S.~Park, and J.~Kim.
\newblock An ert-based robotic skin with sparsely distributed electrodes:
  Structure, fabrication, and dnn-based signal processing.
\newblock In {\em 2020 IEEE International Conference on Robotics and Automation
  (ICRA)}, pages 1617--1624, 2020.

\bibitem{Chen:2021}
Y.~Chen and H.~Liu.
\newblock Location-dependent performance of large-area piezoresistive tactile
  sensors based on electrical impedance tomography.
\newblock {\em IEEE Sensors Journal}, 21(19):21622--21630, 2021.

\bibitem{Park:2021}
H.~Park, K.~Park, S.~Mo, and J.~Kim.
\newblock Deep neural network based electrical impedance tomographic sensing
  methodology for large-area robotic tactile sensing.
\newblock {\em IEEE Transactions on Robotics}, 37(5):1570--1583, 2021.

\bibitem{Chen:2022}
H.~Chen, X.~Yang, P.~Wang, J.~Geng, G.~Ma, and X.~Wang.
\newblock A large-area flexible tactile sensor for multi-touch and force
  detection using electrical impedance tomography.
\newblock {\em IEEE Sensors Journal}, 22(7):7119--7129, 2022.

\bibitem{Park:2022}
K.~Park, H.~Lee, K.~J. Kuchenbecker, and J.~Kim.
\newblock Adaptive optimal measurement algorithm for ert-based large-area
  tactile sensors.
\newblock {\em IEEE/ASME Transactions on Mechatronics}, 27(1):304--314, 2022.

\bibitem{Rus:2015}
D.~Rus and M.~T. Tolley.
\newblock Design, fabrication and control of soft robots.
\newblock {\em Nature}, 521(7553):467--475, 2015.

\bibitem{Yoon:2017}
S.~Ho Yoon, K.~Huo, Y.~Zhang, G.~Chen, L.~Paredes, S.~Chidambaram, and K.~Ramani.
\newblock Isoft: A customizable soft sensor with real-time continuous contact
  and stretching sensing.
\newblock In {\em Proceedings of the 30th Annual ACM Symposium on User
  Interface Software and Technology}, UIST '17, pages 665--678, New York, NY,
  USA, 2017. Association for Computing Machinery.

\bibitem{Park:2022_s}
K.~Park, H.~Yuk, M.~Yang, J.~Cho, H.~Lee, and J.~Kim.
\newblock A biomimetic elastomeric robot skin using electrical impedance and
  acoustic tomography for tactile sensing.
\newblock {\em Science Robotics}, 7(67):eabm7187, 2022.

\bibitem{Preti:2022}
M.~Lo Preti, M.~Totaro, E.~Falotico, M.~Crepaldi, and L.~Beccai.
\newblock Online pressure map reconstruction in a multitouch soft optical
  waveguide skin.
\newblock {\em IEEE/ASME Transactions on Mechatronics}, pages 1--11, 2022.

\bibitem{Marashdeh:2015}
Q.~Marashdeh, F.~L. Teixeira, and L.~S. Fan.
\newblock {\em 1 - Electrical capacitance tomography}, pages 3--21.
\newblock Woodhead Publishing, 2015.

\bibitem{Pagoli:2022vq}
A.~Pagoli, F.~Chapelle, J.~Corrales-Ramon, .~Mezouar, and Y.~Lapusta.
\newblock Large-area and low-cost force/tactile capacitive sensor for soft
  robotic applications.
\newblock {\em Sensors}, 22(11), 2022.

\bibitem{Yang:2003}
W.~Q Yang and L.~Peng.
\newblock Image reconstruction algorithms for electrical capacitance
  tomography.
\newblock {\em Measurement Science and Technology}, 14(1):R1, 2003.

\bibitem{YangY:2017}
Y.~Yang, L.~Peng, and J.~Jia.
\newblock A novel multi-electrode sensing strategy for electrical capacitance
  tomography with ultra-low dynamic range.
\newblock {\em Flow Measurement and Instrumentation}, 53:67--79, 2017.

\bibitem{LeCun:2015}
Y.~LeCun, Y.~Bengio, and G.~Hinton.
\newblock Deep learning.
\newblock {\em Nature}, 521(7553):436--444, 2015.

\bibitem{Kingma:2015}
D.~Kingma and J.~Ba.
\newblock Adam: A method for stochastic optimization.
\newblock In {\em International Conference on Learning Representations (ICLR)},
  2015.

\bibitem{Delin:2022}
D.~Hu, F.~Giorgio-Serchi, S.~Zhang, and Y.~Yang.
\newblock Stretchable e-skin and transformer enable high-resolution
  morphological reconstruction for soft robots.
\newblock {\em Nature Machine Intelligence}, 5(3):261--272, 2023.

\end{thebibliography}
\end{document}